\documentclass[twoside]{article}

\usepackage[accepted]{aistats2023}

\usepackage[utf8]{inputenc} 
\usepackage[T1]{fontenc}    
\usepackage{hyperref}       
\usepackage{url}            
\usepackage{booktabs}       
\usepackage{amsfonts}       
\usepackage{nicefrac}       
\usepackage{microtype}      
\usepackage[dvipsnames]{xcolor}
\usepackage{amsmath}
\usepackage{svg}
\usepackage{tikz}
\usetikzlibrary{calc}
\usepackage{xfrac}
\usepackage{subfigure}

\usepackage{graphicx}
\usepackage{pifont}
\usepackage{multirow}
\usepackage{booktabs}
\usepackage{adjustbox}
\usepackage{graphicx}
\usepackage{mathtools}
\usepackage{amsfonts}
\usepackage{bm}

 \newcommand{\KL}[2]{\mathrm{KL}\left[ {#1} \,\Vert\, {#2} \right]}

\newcommand{\inv}{^{-1}}

\renewcommand{\vec}[1]{{\boldsymbol{\mathrm{#1}}}}

\newcommand{\Kuu}{\vec{K}_{\vec{uu}}}
\newcommand{\Kuf}{\vec{K}_{\vec{uf}}}
\newcommand{\Kfu}{\vec{K}_{\vec{fu}}}

\usepackage[english]{babel}

\usepackage{amsthm}

\theoremstyle{definition}

\theoremstyle{definition}
\newtheorem{defn}{Definition}

\theoremstyle{remark}
\newtheorem{remark}{Remark}

\usepackage{enumitem}
\setenumerate{leftmargin=3.5mm}
\setitemize{leftmargin=3.5mm}

\def\diff{\mathrm{d}}
\def\T{\top}

\usepackage[round]{natbib}
\begin{document}

%

%
\runningauthor{Harry Jake Cunningham, Daniel Augusto de Souza, So Takao, Mark van der Wilk, Marc Peter Deisenroth}

\twocolumn[

\aistatstitle{Actually Sparse Variational Gaussian Processes}

\aistatsauthor{Harry Jake Cunningham$\textsuperscript{1}$ \And Daniel Augusto de Souza$\textsuperscript{1}$ \And So Takao$\textsuperscript{1}$}
\aistatsauthor{Mark van der Wilk$\textsuperscript{2}$ \And Marc Peter Deisenroth$\textsuperscript{1}$ }

\aistatsaddress{ University College London$\textsuperscript{1}$, Imperial College London$\textsuperscript{2}$} ]

\begin{abstract}
    Gaussian processes (GPs) are typically criticised for their unfavourable scaling in both computational and memory requirements. For large datasets, sparse GPs reduce these demands by conditioning on a small set of inducing variables designed to summarise the data. In practice however, for large datasets requiring many inducing variables, such as low-lengthscale spatial data, even sparse GPs can become computationally expensive, limited by the number of inducing variables one can use. In this work, we propose a new class of inter-domain variational GP, constructed by projecting a GP onto a set of compactly supported B-spline basis functions. The key benefit of our approach is that the compact support of the B-spline basis functions admits the use of sparse linear algebra to significantly speed up matrix operations and drastically reduce the memory footprint. This allows us to very efficiently model fast-varying spatial phenomena with tens of thousands of inducing variables, where previous approaches failed.
\end{abstract}

\section{INTRODUCTION}

\looseness=-1 Gaussian processes (GPs) \citep{williams2006gaussian} provide a rich prior over functions. Their non-parametric form, gold-standard uncertainty estimates and robustness to overfitting have made them common place in geostatistics \citep{oliver1990kriging}, epidemiology \citep{Bhatt2017}, spatio-temporal modelling \citep{blangiardo2013spatial, wikle2019spatio}, robotics and control \citep{deisenroth2011pilco} and Bayesian optimisation \citep{osborne2009gaussian}.  However, GPs scale infamously as $\mathcal{O}(N^3)$ in computational complexity and $\mathcal{O}(N^2)$ in memory, where $N$ is the size of the training dataset, making them unfeasible for use with large datasets. To overcome this limitation, there exist a number of different approximate inference techniques, including sparse approximations \citep{snelson2006sparse, quinonero2005unifying, titsias2009variational}, state-space methods \citep{hartikainen2010kalman, sarkka2013spatiotemporal, hamelijnck2021spatio} and local-expert models \citep{tresp2000bayesian, tresp2000mixtures, rasmussen2001infinite, deisenroth2015distributed, Cohen2020}.
In particular, sparse GP approximations have been developed to reduce the cubic complexity of inference by introducing a set of inducing variables. Sparse approaches summarise the training data by a set of $M\ll N$ pseudo-data, effectively reducing the rank of the covariance matrix. Amongst these methods, variational approximations have proved popular in improving GPs for regression \citep{titsias2009variational}, classification \citep{hensman2015scalable}, stochastic optimisation \citep{hensman2013gaussian}, inference with non-conjugate likelihoods \citep{hensman2015scalable, hensman2015mcmc} and hierarchical non-parametric modelling \citep{damianou2013deep, salimbeni2017doubly}. 

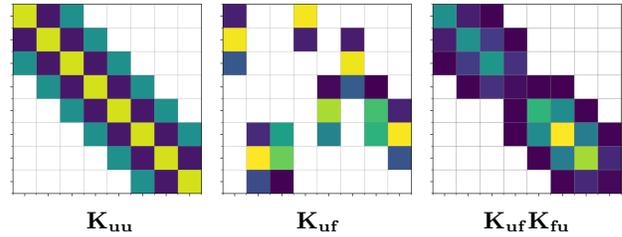
\begin{figure}[t]
\begin{tikzpicture}
\node [
    above right,
    inner sep=0] (image) at (0,0) {\includegraphics[width=\columnwidth]{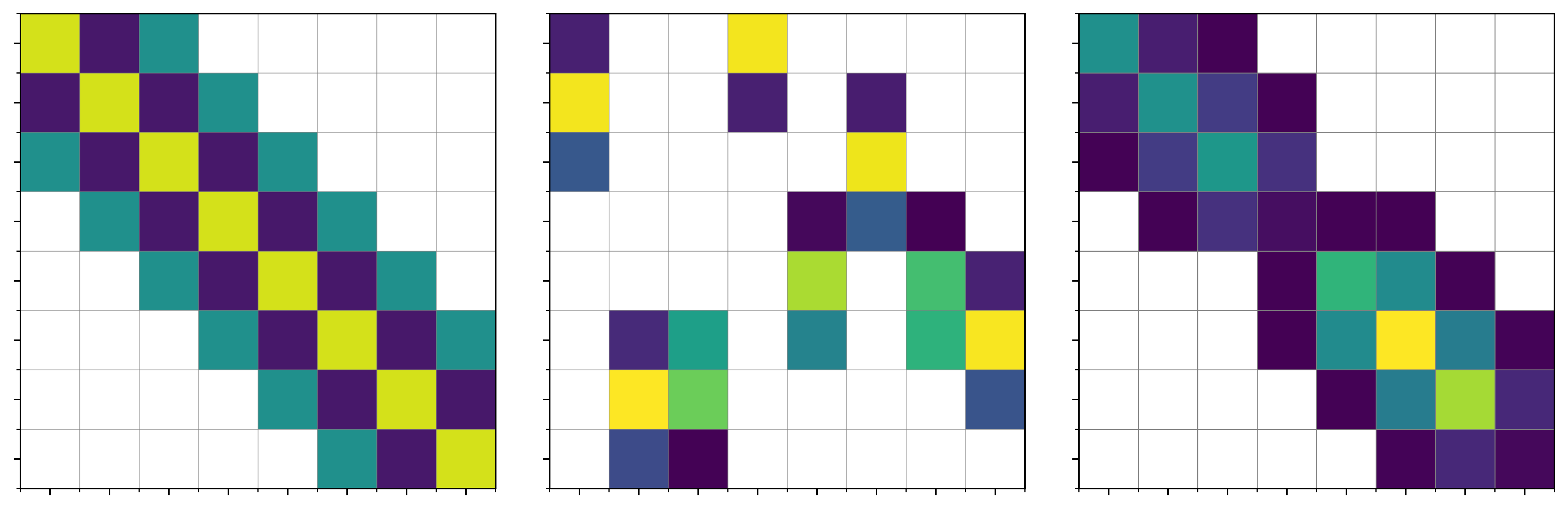}};
\begin{scope}[
x={($0.1*(image.south east)$)},
y={($0.1*(image.north west)$)}] 
    \node[black,fill=white] at (1.7, -1) {\small $\mathbf{K_{uu}}$};
    \node[black,fill=white] at (5.05, -1) {\small $\mathbf{K_{uf}}$};
    \node[black,fill=white] at (8.4, -1) {\small $\mathbf{K_{uf}}\mathbf{K_{fu}}$};
\end{scope}
\end{tikzpicture}
\vspace*{-6mm}
\caption{Illustration of the sparse matrix structures induced by our proposed method for 1D regression with a Mat\'ern-3/2 kernel. By constructing inter-domain inducing variables $\mathbf{u}$ as RKHS projections of the GP onto a set of compactly supported B-splines, both the inducing point covariance matrix $\mathbf{K_{uu}}$ and the covariance matrix between the GP $f$ and the inducing variables $\mathbf{K_{uf}}$ become sparse. This admits sparse linear algebra to precompute the sparse matrix product $\mathbf{K_{uf}}\mathbf{K_{fu}}$, which is used to compute the ELBO.}
\label{fig:matrices}
\end{figure}

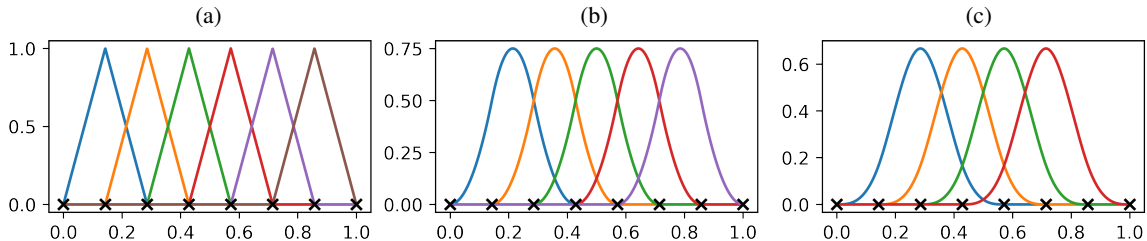
\begin{figure*}[t]
\centering
\begin{tikzpicture}
\node [
    above right,
    inner sep=0] (image) at (0,0) {\includegraphics[width=0.9\textwidth]{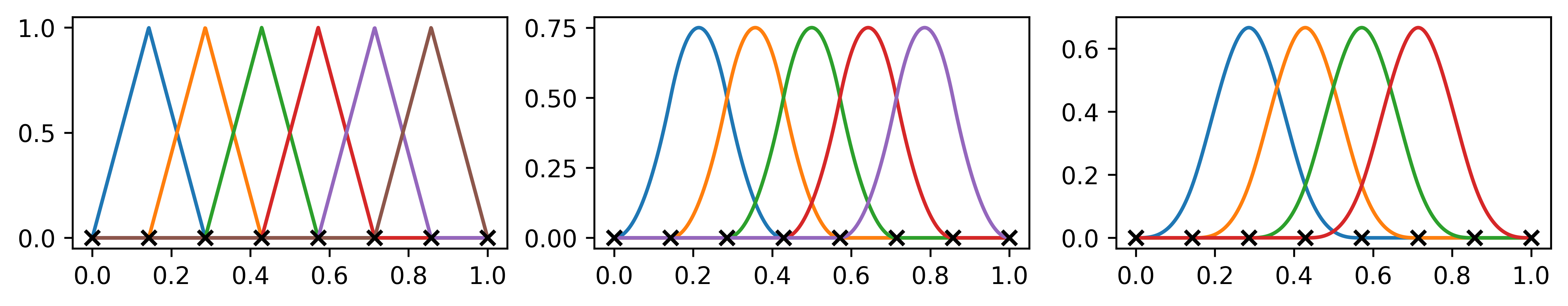}};
\begin{scope}[
x={($0.1*(image.south east)$)},
y={($0.1*(image.north west)$)}] 
    \node[black,fill=white] at (1.84, 10.5) {\small (a)};
    \node[black,fill=white] at (5.16, 10.5) {\small (b)};
    \node[black,fill=white] at (8.47, 10.5) {\small (c)};
\end{scope}
\end{tikzpicture}
\vspace*{-4mm}
\caption{(a) 1st-order B-spline basis (b) 2nd-order B-spline basis (c) 3rd-order B-spline basis. For the same set of knots, the support of the B-splines increases in width with increasing order. This has the effect that each B-spline basis function has intersecting support with an increasing number of basis functions as the order increases.}
\label{fig:splines}
\end{figure*}

Introduced by \cite{titsias2009variational}, Sparse Variational Gaussian processes (SVGPs) approximate the true GP posterior with an approximate one, conditioned on a set of $M$ inducing variables. The approximate posterior is then learnt by minimising the Kullback-Leibler (KL) divergence between the approximate and true posterior, allowing us to learn the variational parameters and hyperparameters jointly via gradient descent. The resulting approximation scales as $\mathcal{O}(NM^2 + M^3)$ in computational complexity and $\mathcal{O}(NM)$ in memory. However, these low-rank approximations are practically limited to $\approx 10,000$ inducing points, which can be insufficient for complex datasets where a large number of inducing points are required to cover the input space. This limitation is especially apparent in long time-series or spatial datasets with intrinsically low lengthscales, where traditional low-rank approximations based on small sets of localised pseudo-datapoints fail to capture fast variations in the data \citep{pleiss2020fast, wu2022variational}.

\looseness=-1To alleviate some of these problems, inter-domain GPs \citep{lazaro2009inter, van2020framework} generalise the idea of inducing variables by transforming the GP to a different domain by means of a linear operator, which admits more expressive features and/or computationally efficient linear algebra. Variational Fourier Features (VFFs) \citep{hensman2017variational}, constructs inter-domain inducing variables by projecting the GP onto a Fourier basis. This results in inducing variables that span the width of the domain and therefore describe global variations in the data. By the orthogonality of the Fourier basis, the inducing variables are also almost independent, producing computationally efficient block-diagonal covariance matrices. In one dimension, this can be exploited to reduce the computational complexity to $\mathcal{O}(M^3)$ after an initial one-off pre-computation of $\mathcal{O}(NM^2)$. 
However, since the Fourier basis functions are global, whilst computationally efficient, they are inefficient at modelling low-lengthscale data. Indeed, VFF typically requires more inducing variables for an equivalent accuracy than standard sparse GP regression for $d\ge 2$ \citep{hensman2017variational}.

Variational Inducing Spherical Harmonics (VISH) by \citet{dutordoir2020sparse} remedied some of the problems faced by VFF by first projecting the data onto a $D$-dimensional unit hypersphere and then using a basis of spherical harmonics as inter-domain inducing features. As the basis functions are orthogonal, VISH reduces the cost of matrix inversion to $\mathcal{O}(M)$ and the total cost of inference to $\mathcal{O}(N M^2)$. However, by projecting data onto the hypersphere and performing sparse GP regression on the transformed space, VISH is unable to use covariance functions which use the Euclidean distance between data points. This makes VISH sub-optimal for naturally Euclidean spatial data. 

In this work, we propose a new inter-domain approach that scales GPs to complex datasets that require a very large number of inducing variables. Specifically, we define a new inter-domain approximation by projecting the GP onto a basis of compactly supported B-splines. 
Due to the local support of the B-spline basis functions, the covariance between inducing variables yields sparse band-diagonal covariance matrices, admitting highly efficient sparse linear algebra at a complexity that scales linearly with the number of inducing variables. In contrast to both VFF and VISH, which use basis functions with global support, our choice of basis also incites sparse structure in the covariance between inducing variables and the GP itself. 
Our results show that our method is particularly well suited to spatial data with high-frequency variations, which necessitate a large number of inducing variables. By using computationally cheap, locally supported inducing variables, we can cover the domain with many basis functions that are able to successfully capture local variations.

\section{BACKGROUND}

\looseness=-1 A Gaussian process is a collection of random variables, any finite number of which is jointly Gaussian distributed. A GP is fully characterised by its mean $\mu(\cdot)$ and covariance function $k(\cdot,\cdot)$ \citep{williams2006gaussian}.
Given a training dataset $\mathcal{D} = \{(\mathbf{x}_n, y_n)\}^N_{n=1}$ of $N$ noisy observations $y_n\in\mathbb{R}$ and corresponding inputs $\mathbf{x}_n\in\mathbb{R}^D$, and observation model $y_n = f(\mathbf{x}_n) + \epsilon,~\epsilon\sim\mathcal N(0, \sigma^2)$, we construct a GP regression problem by placing a zero-mean GP prior on the latent function $f\sim\mathcal{GP}(0, k(\cdot,\cdot))$. 
The posterior distribution $p(f|\mathbf{y})\sim\mathcal{GP}(\mu(\cdot), \Sigma(\cdot,\cdot))$ is a GP with
\begin{align}
\begin{aligned}
    \mu(\cdot) &= \mathbf{k}_{\mathbf f}^T(\cdot)\mathbf{K}_{\mathbf{yy}}^{-1}\mathbf{y}, \\
    \Sigma(\cdot,\cdot) &= k(\cdot,\cdot) - \mathbf{k}_{\mathbf f}^T(\cdot)\mathbf{K}_{\mathbf{yy}}^{-1}\mathbf{k_f}(\cdot),
    \label{eq:gp posterior}
\end{aligned}
\end{align}
where $\mathbf{k_f}(\cdot) = [k(\mathbf{x}_n, \cdot)]^N_{n=1}$, $\mathbf{K_{yy}} = \mathbf{K_{ff}}+\sigma^2\mathbf{I}$ and $\mathbf{K_{ff}}=[k(\mathbf{x}_i, \mathbf{x}_j)]^N_{i,j=1}$.

To train the GP we maximise the log-marginal likelihood $\log p(\vec{y}) = \log \int p(\vec{y} | \vec{f}) p(\vec{f}) \diff \vec{f}$. In the case of a Gaussian likelihood, this takes the explicit form
\begin{align} \label{eq:log-evidence}
    \log p(\vec{y}) = -\frac12 \vec{y}^\T \mathbf{K}_{\vec{yy}}^{-1} \vec{y} - \frac12\log |\mathbf{K_{yy}}| - \frac{n}{2} \log{2\pi}.
\end{align}
%
\looseness=-1 Training the GP scales in $\mathcal O(N^3)$ due to computing the matrix inverse and determinant in \eqref{eq:log-evidence}. Moreover, when using gradient-based optimisation to tune the hyperparameters,  \eqref{eq:log-evidence} must be computed at every iteration. Predictions using \eqref{eq:gp posterior} require $\mathcal O(N^2)$ computations, assuming $\mathbf{K}_{\mathbf{yy}}\inv$ (or its Cholesky factorisation) has been cached, e.g., after the training procedure. In terms of memory, GP predictions require $\mathcal O(N^2)$ to store the Cholesky factor of $\mathbf{K}_{\mathbf{yy}}$. The computational and memory demands therefore make GPs prohibitively expensive for datasets with more than $\approx 10,000$ datapoints.

\subsection{Sparse Variational Gaussian Processes}

Variational inference provides an elegant method to approximate the true posterior $p(f|\mathbf{y})$ of a GP with a variational distribution $q(f)$, rather than approximating the model itself. Sparse variational Gaussian processes (SVGPs) introduced by \cite{titsias2009variational} leverage inducing points coupled with variational inference to construct a low-rank approximation to the posterior. SVGP consists of introducing a (small) set of inducing variables $\mathbf{u}=\{f(\mathbf{z}_m)\}^M_{m=1}$ defined at a set of inducing point locations $Z=\{\mathbf{z}_m\}^M_{m=1}$. Placing a Gaussian distribution over the inducing variables $q(\mathbf{u})=\mathcal{N}(\mathbf{m},\mathbf{S})$, the approximate posterior
\begin{equation}
    q(f) = \int p(f|\mathbf{u})q(\mathbf{u})\diff\mathbf{u} = \mathcal{GP}(\mu(\cdot), \Sigma(\cdot, \cdot))
    \label{eq:approx post}
\end{equation}
is obtained by marginalising out the inducing variables. The approximate posterior \eqref{eq:approx post} is defined in terms of the variational parameters $\mathbf{m}\in\mathbb{R}^M$ and $\mathbf{S}\in\mathbb{R}^{M\times M}$,
where, due to the conjugacy between $p(f|\mathbf{u})$ and $q(\mathbf{u})$,
\begin{align}
    \mu(\cdot) &= \mathbf{k}_{\mathbf{u}}^T(\cdot)\mathbf{K}_{\mathbf{uu}}^{-1}\mathbf{m}, \label{eq:q-mean} \\
    \Sigma(\cdot,\cdot) &= k(\cdot,\cdot)+\mathbf{k}_{\mathbf{u}}^T(\cdot)\mathbf{K}_{\mathbf{uu}}^{-1}(\mathbf{S}-\mathbf{K_{uu}})\mathbf{K}_{\mathbf{uu}}^{-1}\mathbf{k_u}(\cdot). \label{eq:q-cov}
\end{align}
Here $\mathbf{k_u}(\cdot)=[\text{cov}({u}_m,f(\cdot))]^M_{m=1}=[k(\mathbf{z}_m, \cdot)]^M_{m=1}$ and $\mathbf{K_{uu}}=[\text{cov}({u}_i, {u}_j)]_{i,j=1}^M=[k(\mathbf{z}_i, \mathbf{z}_j)]_{i,j=1}^M$.

The variational parameters $\mathbf{m}$ and $\mathbf{S}$ are optimised by minimising the KL divergence between the true and approximate posterior $\KL{q(f)}{p(f|\mathbf{y})}$. In practice, this is made tractable by maximising the evidence lower bound (ELBO)
\begin{equation}
    \mathcal{L}_{\mathrm{ELBO}} = \sum^N_{n=1} \mathbb{E}_{q(f_n)}[\log p(y_n|f_n)] - \KL{q(\mathbf{u})}{p(\mathbf{u})},
    \label{Eq:SVGP_ELBO}
\end{equation}
which provides a lower bound to the log-marginal likelihood $\log{p(\mathbf{y})} \ge \mathcal{L}_{\mathrm{ELBO}}$, and whose gap is precisely the KL divergence that we are minimising. Normally, the hyperparameters of the model are optimised jointly with the variational parameters, by maximising the ELBO.

For a Gaussian likelihood, the moments of the optimal distribution $\hat{q}(\vec{u}) = \mathcal{N}(\hat{\vec{m}}, \hat{\vec{\Sigma}})$ can be computed exactly as
\begin{align}
    \hat{\vec{m}} &= \sigma^{-2} \hat{\vec{\Sigma}} \Kuf \vec{y}, \label{eq:optimal-mean} \\
    \hat{\vec{\Sigma}} &= \Kuu \left[\Kuu + \sigma^{-2} \Kuf \Kfu\right]^{-1} \Kuu. \label{eq:optimal-cov}
\end{align}
The corresponding optimal ELBO is given by
\begin{align} \label{eq:ELBO}
\begin{aligned}
    \mathcal{L}_{\mathrm{ELBO}} &=  \log \mathcal{N}\left(\mathbf{y} | \mathbf{0}, \Kfu\Kuu^{-1}\Kuf + \sigma_n^2\mathbf{I}\right) \\
    &\quad- \frac{1}{2}\sigma_n^{-2}\text{tr}\left(\mathbf{K_{ff}} - \Kfu\Kuu^{-1}\Kuf\right),
    \end{aligned}
\end{align}
where $\mathbf{K_{uf}}=[k(\mathbf{z}_m, \mathbf{x}_n)]^{M,N}_{m,n=1}$.
SVGPs thus reduce the computational cost of training to $\mathcal{O}(NM^2 + M^3)$ per evaluation of the ELBO. \cite{hensman2013gaussian} showed that the ELBO in \eqref{Eq:SVGP_ELBO} is also amenable to stochastic optimisation, further reducing the computational complexity to $\mathcal{O}(N_b M^2 + M^3)$ per iteration by using minibatches. 
SVGPs require $\mathcal O(N_bM + M^2)$ memory to store $\mathbf{K_{fu}}$ and the dense Cholesky factor of $\mathbf{K_{uu}}$.

The use of a low-rank approximation does have certain trade-offs, however. Whilst small $M$ speeds up computation, the choice of $M$ is also essential to ensuring a certain quality of approximation \citep{burt2019rates}. Using a small number of inducing points becomes particularly troublesome for data with inherently short lengthscales, which commonly occurs when working with spatial data. In this case, the SVGP will collapse quickly to the prior mean and variance when not in the immediate vicinity of an inducing input.

\subsection{Variational Fourier Features (VFF)} 

Inter-domain GPs \citep{alvarez2008sparse, lazaro2009inter, van2020framework} generalise the idea of inducing variables by instead conditioning on a linear transformation $\mathcal{L}_m$ of the GP $\mathbf{u} = [ \mathcal{L}_mf(\cdot)]^M_{m=1}$. 
By choosing $\mathcal{L}_m$ to be a convolution of $f(\cdot)$ with respect to a Dirac delta function centred at the inducing points $\vec{z}_m$, we can recover the standard inducing point approximation. 
However, by choosing different linear operators, such as projections \citep{hensman2017variational, dutordoir2020sparse} or general convolutions \citep{van2017convolutional}, we can construct more informative features, without changing the sparse variational inference scheme.

VFF \citep{hensman2017variational} is an inter-domain variational GP approximation that constructs inducing features
as a Mat\'ern RKHS projection of the GP onto a set of Fourier basis functions
$
    u_m = \langle f, \phi_m \rangle_{\mathcal{H}}, m=1, \ldots, M,
$
where $\left<\cdot, \cdot\right>_{\mathcal{H}}$ denotes the Mat\'ern RKHS inner product, and $\phi_0(x)=1$, $\phi_{2i-1}(x)=\cos(\omega_i x)$, $\phi_{2i} = \sin(\omega_i x)$ are the Fourier basis functions. This results in the matrices
\begin{equation}
        \Kuu = [\langle \phi_i, \phi_j \rangle_{\mathcal{H}}]_{i,j=1}^{M}, \quad \Kuf = [\phi_m(x_n)]_{m, n=1}^{M,N}
\end{equation}
\looseness=-1where, due to the reproducing property, 
the cross-covariance matrix $\Kuf$,
which is equivalent to evaluating the Fourier basis, is independent of kernel hyperparameters. This leads to several computational benefits: (i) we can precompute
$\Kuf$, 
as it remains constant throughout hyper-parameter training via the ELBO \eqref{eq:ELBO}, (ii) due to the orthogonality of the Fourier basis, $\Kuu$ is the sum of a block-diagonal matrix plus low-rank matrices, e.g., in the case of a 1D Matérn-1/2 kernel, 
\begin{align}
    \Kuu =  \mathrm{diag}(\boldsymbol{\alpha}) + \boldsymbol{\beta} \boldsymbol{\beta}^\T
\end{align}
for some $\boldsymbol{\alpha}, \boldsymbol{\beta} \in \mathbb{R}^M$, where the vector $\boldsymbol{\beta}$ is sparse.
This structure can be exploited to significantly reduce the computational complexity for training and prediction when compared to standard sparse GP methods. 
However, VFF has two main flaws:
\begin{itemize}
\item \looseness=-1 VFF generalises poorly to higher dimensions due to the use of a Kronecker product basis. This construction of a high-dimensional basis not only scales exponentially in the number of dimensions, it is also inefficient in terms of captured variance \citep{dutordoir2020sparse}: Multiplying together basis functions of increasing frequency causes the prior variance to decay rapidly, resulting in large numbers of redundant features and the down-weighting of important low-frequency ones. Thus for $D\ge 2$, VFF typically requires more inducing variables than SGPR, making it memory inefficient.
\item Whilst $\Kuu$ has a computationally efficient structure, $\Kuf$ is still a dense matrix. In the special case when the likelihood is Gaussian, we still require to compute a dense Cholesky factor of the $M\times M$ matrix $\Kuu + \sigma^{-2}\Kuf\Kfu$ (see \eqref{eq:optimal-cov}), which costs $\mathcal{O}(M^3)$. 
The same problem persists for VISH.
\end{itemize}
In order to address these issues, in the next section we will consider defining inter-domain inducing variables as the projection of the GP onto a set of compactly supported basis functions, drastically reducing memory requirements and improving computational efficiency, enabling us to use large numbers of inducing points.

\section{B-SPLINE INDUCING FEATURES}

In this section, we introduce B-spline inducing features and propose Actually Sparse Variational Gaussian Processes (AS-VGPs). The core idea is to use the concept of RKHS projections as in VFF, except to project a GP onto a set of compactly supported {\em B-spline basis functions} instead of the Fourier basis functions. Unlike in VFF, the resulting inducing features $\{u_m\}_{m=1}^M$ are localised by the nature of their compact support, see Figure \ref{fig:splines}, such that $\Kuu$, $\Kuf$ and $\Kuf\Kfu$ are $\emph{all}$ sparse matrices (see Figure \ref{fig:matrices}). These sparse covariance structures allow us to gain substantial computational benefits.

\subsection{B-Spline Inducing Features} 

B-spline basis functions of order $k$ are a set of compactly supported piece-wise polynomial functions of degree $k$. Their shape is controlled by an increasing sequence of knots $V=\{v_m\}^M_{m=0}\in\mathbb{R}$ that partition the domain into $M$ sub-intervals. We denote the $m$-th B-spline basis function of order $k$ by $B_{m,k}(x)$ (See Appendix \ref{app:splines} for expressions). Since a $k$-th order B-spline has compact support over only $k+1$ sub-intervals, it has intersecting support with at most $k+1$ other B-spline basis functions (see Figure \ref{fig:splines}). 

We define the {\em B-spline inducing features} as the RKHS projection $u_m = \langle f, \phi_m(\cdot) \rangle_{\mathcal{H}}$ onto the B-spline basis, where $\phi_m(x) = B_{m,k}(x)$.
Under this choice, the covariance between the inducing features $u_m$ and the GP $f$ is given by
\begin{align}
    [\mathbf{K_{uf}}]_{m,n} &= \mathrm{Cov}[u_m, f(x_n)] = \langle k(x_n,\cdot), \phi_m(\cdot) \rangle_\mathcal{H} \\
    &= \phi_m(x_n) = B_{m,k}(x_n) \label{eq:ku-B-spline}
\end{align}
and reduces to a simple evaluation of the B-spline basis at the training inputs.
Note that $B_{m,k}(x)\neq 0$ if and only if $x\in [v_m, v_{m+k+1}]$ and therefore $\Kuf$ is sparse with at most $M(k+1)$ non-zero entries. As with VFF, $\Kuf$ is also independent of the kernel hyperparameters, meaning it remains constant throughout training and can be precomputed. 
Next, the covariance between the inducing features is given by
\begin{equation} \label{eq:Kuu}
    [\Kuu]_{m,m'} = \mathrm{Cov}[u_m, u_{m'}] = \langle \phi_m, \phi_{m'} \rangle_{\mathcal{H}},
\end{equation}
which is only non-zero when $\phi_m$ and $\phi_{m'}$ have intersecting support. This produces sparse band-diagonal $\Kuu$ matrices with bandwidth equal to $k+1$. Since the B-spline basis functions are piecewise polynomials, we can evaluate the inner product in closed form, allowing for efficient computation during training and testing.

\begin{remark}
Strictly speaking, the notation $\langle f, \phi_m(\cdot) \rangle_{\mathcal{H}}$ is ill-defined, as samples of $f$ are almost surely not elements of $\mathcal{H}$ \cite{kanagawa2018gaussian}. In order to make rigorous sense of this, we use the machinery of generalised Gaussian fields, which we discuss in Appendix \ref{app:generalised-gaussian-fields}.
\end{remark}

\subsection{Sparse Linear Algebra} \label{Actually Sparse Variational GP Inference}


\begin{table*}
  \caption{Complexity of sparse variational GPs for evaluating the ELBO \eqref{eq:ELBO} in 1D regression settings with a Gaussian likelihood. $N$: number of datapoints; $M$: number of inducing points; $k$: bandwidth of the covariance matrix; $N_b$: size of the mini-batch in stochastic variational inference. For both VFF and VISH we quote the complexity required for exact SGPR. \\}
  \label{Tab:SGPR_Complexity}
  \centering
  \scalebox{1}{
  \begin{tabular}{llll}
    \toprule
    Algorithm & Pre- & Computational  & Storage \\
            &    computation               & complexity &\\
    \midrule
    SGPR \citep{titsias2009variational} & \ding{55} & $\mathcal{O}(NM^2+M^3)$  & $\mathcal{O}(NM)$    \\
    SVGP \citep{hensman2013gaussian} & \ding{55} & {$\mathcal{O}(N_b M^2 + M^3)$} & {$\mathcal{O}(M^2 + N_bM)$ } \\
    VFF \citep{hensman2017variational} & $\mathcal{O}(NM^2)$ & {$\mathcal{O}(M^3)$} & {$\mathcal{O}(NM + M^2)$ }    \\
    VISH \citep{dutordoir2020sparse} & $\mathcal{O}(NM^2)$ & {$\mathcal{O}(M^3)$} & {$\mathcal{O}(NM + M^2)$} \\
    \textbf{AS-VGP (Ours)} & \textcolor{Green}{$\bm{\mathcal{O}(N)}$} & \textcolor{Green}{$\bm{\mathcal{O}((k+1)^2M)}$}  & \textcolor{Green}{$\bm{\mathcal{O}(N + (k+1)M)}$}  \\
    \bottomrule
  \end{tabular}
  }
\end{table*}

In this section, we will initially restrict our analysis to GPs with one-dimensional inputs and extend this later in Section \ref{sec:higher-dimensions} to higher dimensions. Using our proposed spline inducing features, we have the following desirable properties that we can leverage in the key computations \eqref{eq:optimal-cov}--\eqref{eq:ELBO}:

\textbf{Property 1:} \quad For the Mat\'ern-$\nu/2$ class of kernels, $\mathbf{K_{uu}}$ is a band-diagonal matrix with bandwidth equal to \textit{at least} $\nu/2+3/2$.

This is due to the fact that, in order to be a valid projection, the B-spline basis functions must belong to the same Mat\'ern RKHS. As stated by   \cite{kanagawa2018gaussian}, the RKHS generated by the Mat\'ern-$\nu/2$ kernel $k(\cdot, \cdot)$ is norm-equivalent to the Sobolev space $\mathcal{H}^{\nu/2+1/2}$. Given their polynomial form, we can check that B-splines of order $k$ are $C^{k
-1}$-smooth and moreover $k$-times weakly differentiable (see Appendix \ref{app:splines}). Since the B-splines are compactly supported, so are their (weak) derivatives;   therefore, the (weak) derivatives are all square-integrable. Thus, they belong to the Sobolev space $\mathcal{H}^{k}$. As a result, for the Mat\'ern-$\nu/2$ kernel, we choose to project onto B-splines of order $k=\nu/2+1/2$, giving us a $\Kuu$ matrix with bandwidth $k+1 = \nu/2+3/2$.

\textbf{Property 2:} The matrix product $\mathbf{K_{uf}}\mathbf{K_{fu}}$ is a band-diagonal matrix with bandwidth at most equal to that of $\Kuu$.

To see this, from \eqref{eq:ku-B-spline} we have
\begin{align}
    [\Kuf\Kfu]_{ij} &= \sum_{n=1}^N [\Kuf]_{in}[\Kuf]_{jn} \\
    &= \sum_{n=1}^N B_{i, k}(x_n) B_{j, k}(x_n). \label{eq:ij-th-element-of-product}
\end{align}
By the properties of B-splines, $B_{i, k}(x_n) B_{j, k}(x_n) \neq 0$ if and only if $x_n \in \mathcal{I}_{ij}$, where $\mathcal{I}_{ij} = [v_i, v_{i+k+1}] \cap [v_j, v_{j+k+1}]$ is the intersection of the supports of the two B-splines $B_{i, k}$ and $B_{j, k}$. However, we know that the supports are intersecting if and only if $|i-j| < k+1$. Hence, when $|i-j| \geq k+1$, no data point can be contained in $\mathcal{I}_{ij}$ since it is the empty set, giving us $B_{i, k}(x_n) B_{j, k}(x_n) = 0$ for all $n=1, \ldots, N$ and therefore $[\Kuf\Kfu]_{ij} = 0$ from \eqref{eq:ij-th-element-of-product}. This implies that the matrix $\Kuf\Kfu$ has bandwidth at most equal to $k+1$.

Using these two properties, we can construct an inter-domain variational method that can leverage sparse linear algebra to speed up inference and significantly save on memory footprint. We discuss this next.

\begin{table*}[ht]
  \caption{Predictive mean squared errors (MSEs) and negative log predictive densities (NLPDs) with one standard deviation based on 5 random splits for a number of UCI regression datasets. All models use a Mat\'ern-3/2 kernel and L-BFGS optimiser.}
  \label{Tab:regression_benchmakrs}
  \centering
  \resizebox{2\columnwidth}{!}{%
  \begin{tabular}{lllllllll}
  \toprule
  & & &
  \multicolumn{3}{c}{MSE ($\times 10^{-1}$)} &
  \multicolumn{3}{c}{NLPD} \\ 
  \cmidrule(lr){4-6}
  \cmidrule(lr){7-9}
  {Dataset} & 
  \hfil $N$ & 
  \hfil $M$ & 
  \hfil SGPR & 
  \hfil VFF & 
  \hfil AS-VGP &
  \hfil SGPR & 
  \hfil VFF & 
  \hfil AS-VGP \\
  \midrule
  Air Quality & 9k & 500  & 6.43 $\pm$ 0.04 & 6.64 $\pm$ 0.04 & 6.68 $\pm$ 0.04 & 1.24 $\pm$ 0.00 & \hfil 1.25 $\pm$ 0.00 & \hfil 1.25 $\pm$ 0.00 \\
  Synthetic   & 10k & 50  & 0.40 $\pm$ 0.00 & 0.39 $\pm$ 0.00 & 0.39 $\pm$ 0.00 & \hfil -0.16 $\pm$ 0.00 &\hfil -0.15 $\pm$ 0.00 &\hfil -0.15 $\pm$ 0.00 \\
  Rainfall    & 43k & 700 & {0.48 $\pm$ 0.00} & 0.83 $\pm$ 0.00 & 0.84 $\pm$ 0.00 & \hfil {0.10 $\pm$ 0.00} & \hfil 0.25 $\pm$ 0.00 & \hfil 0.29 $\pm$ 0.00 \\
  Traffic & 48k & 300 & 9.96 $\pm$ 0.01 & 10.01 $\pm$ 0.01 & 10.02 $\pm$ 0.01 & 1.42 $\pm$ 0.00 & 1.42 $\pm$ 0.00 & 1.42 $\pm$ 0.00 \\
  \bottomrule
  \end{tabular}%
  }
\end{table*}

\subsection{Actually Sparse Variational Gaussian Processes}

We propose Actually Sparse Variational Gaussian Processes (AS-VGP) as inter-domain variational GPs that use B-Spline inducing variables.
For one-dimensional GPs, our method has several computational advantages: 
\begin{itemize}
\item $\Kuf$ is very sparse with typically 1\% of its entries being non-zero. This allows us to store it as a sparse tensor, resulting in 2 orders of magnitude memory saving.
\item By Properties (1)--(2), the sum $\Kuu +\sigma^{-1}\Kuf\Kfu$ in \eqref{eq:optimal-cov} is band-diagonal and its inverse can be computed at a cost of $\mathcal O(M(k+1)^2)$; its memory footprint is $\mathcal O(M(k+1))$.
\item Using the banded operators from \cite{durrande2019banded}, we compute $\mathrm{tr}(\Kfu\Kuu^{-1}\Kuf) = \mathrm{tr}(\Kuu^{-1}\Kuf\Kfu)$ in~\eqref{eq:ELBO} without having to instantiate a dense matrix, reducing the memory footprint to $\mathcal O(M(k+1))$.
\end{itemize}

Overall, this reduces the pre-computation cost for computing the sparse matrix multiplication $\Kuf\Kfu$ to \textit{linear} in the number of training datapoints. The resulting matrix can be cached for later use with a memory footprint of $\mathcal{O}((k+1)M)$, owing to its banded structure (Property 2). Further, the per-iteration computational cost and memory footprint of computing the ELBO~\eqref{eq:ELBO} and its gradients is also \textit{linear} in the number of inducing variables, required to take the (sparse) Cholesky decomposition of a banded matrix \citep{durrande2019banded}. 

Further, using the banded operators introduced by \cite{durrande2019banded}, given a banded Cholesky factor of $\Kuu$, we can also compute only the band elements of its inverse at a cost of $\mathcal{O}(M(k+1)^2)$. Given that $\Kuf\Kfu$ is a banded matrix (Property 2), we compute the trace term in \eqref{eq:ELBO} by computing only the bands of the matrix product $\Kuu^{-1}\Kuf\Kfu$, with the computational cost $\mathcal{O}(M(k+1)^2)$, thereby avoiding \emph{ever} instantiating a dense matrix.

We compare the compute and memory costs of various sparse GP inference algorithms in Table \ref{Tab:SGPR_Complexity}. This highlights the linear scaling in both memory and computational complexity with inducing points of the proposed AS-VGP. Compared to both VFF and VISH, AS-VGP is the only method that scales linearly in both computational complexity and storage, enabling it to be used with tens or hundreds of thousands of inducing variables, significantly more than both VFF and VISH. 

\subsection{Extensions to Higher Dimensions}\label{sec:higher-dimensions}

To extend AS-VGP to higher dimensions, we employ a similar strategy to VFF, by constructing either the additive or separable kernel.
In the separable case, we have
\begin{equation}
    k(\mathbf{x}, \mathbf{x}') = \prod^D_{d=1} k_d(x_d, x'_d),
\end{equation}
where $k_d(\cdot, \cdot)$ for $d=1, \ldots, D$ are one-dimensional kernels. By choosing the basis functions to be a tensor product of $M$ one-dimensional B-Splines, that is,
\begin{equation} \label{eq:separable-basis}
    \boldsymbol{\phi}(\vec{x}) = \bigotimes^D_{d=1} [B_{m, k}^{(d)}(x_d)]^M_{m=1} \in \mathbb{R}^{M^D},
\end{equation}
we get the matrices
\begin{equation}
    \Kuf  = [\boldsymbol{\phi}(\vec{x}_n)]_{n=1}^N, \quad \Kuu = \bigotimes_{d=1}^D \Kuu^{(d)},
\end{equation}
computed using \eqref{eq:ku-B-spline}--\eqref{eq:Kuu}, where $\Kuu^{(d)}$ for $d=1, \ldots, D$ denotes the matrix \eqref{eq:Kuu} corresponding to the 1D case and $\{\vec{x}_n\}_{n=1}^N$ are the training inputs.
Note that some of the structures present for one-dimensional inputs are also present in the Kronecker formulation, namely: (i) $\mathbf{K_{uu}}$ is a block-banded matrix with bandwidth $\approx kM^{D-1}$ whose Cholesky factorisation can be computed in $\mathcal{O}(kM^{D-1})$, (ii) $\Kuf\Kfu$ is also a band-diagonal matrix with bandwidth $\approx kM^{D-1}$. For low-dimensional problems, the large number of basis functions \eqref{eq:separable-basis} provides a rich covering of the input space. However, this is unsuitable for large $D$ due to the exponential scaling in the number of input dimensions. 

For the additive case, we construct $D$-dimensional kernels as the sum of $D$ one-dimensional kernels, i.e.,
\begin{equation}
    k(\mathbf{x}, \mathbf{x}) = \sum^D_{d=1} k_d(x_d, x'_d).
\end{equation}
This results in a band-diagonal $\Kuu$ matrix with bandwidth equal to the one-dimensional equivalent. However, the product $\Kuf\Kfu$ is no longer sparse and hence inference using a Gaussian likelihood and pre-computation requires an $\mathcal{O}(DM^3)$ Cholesky factorisation.

\begin{table*}[t]
  \caption{Predictive mean squared errors (MSEs), negative log predictive densities (NLPDs) and wall-clock time in seconds with one standard deviation based on $5$ random splits of the household electric power consumption dataset containing $2,049,279$ data points. The number of inducing variables used is given by $M$. 
  }
  \label{Tab:large_scale_regression}
  \centering
  \scalebox{1}{
  \begin{tabular}{lcccccc}
  \toprule
  Method & \hfil $M = 1000$ & \hfil $M = 5000$ & \hfil $M = 10,000$ & \hfil $M = 20,000$ & \hfil $M = 30,000$\\
  \midrule
  AS-VGP (MSE $\times 10^{-1}$) & {\bf 8.65$\pm$ 0.00} & {\bf 6.55$\pm$ 0.01} & {\bf 4.53 $\pm$ 0.00} & {\bf 3.41$\pm$ 0.01}& {\bf 2.90$\pm$ 0.01} \\
  SVGP (MSE $\times 10^{-1}$) & {9.00 $\pm$ 0.01} & / & / & / & / \\
  \midrule
  AS-VGP (NLPD) & {\bf 1.34 $\pm$ 0.00} & {\bf 1.20 $\pm$ 0.00} & {\bf 1.01 $\pm$ 0.00}  & {\bf 0.86 $\pm$ 0.00} & {\bf 0.77 $\pm$ 0.00}\\
  SVGP (NLPD) & {1.37 $\pm$ 0.00} & / & / & / & / \\
  \midrule
  AS-VGP (Time in s) & {\bf 5.51 $\pm$ 0.10} & {\bf 14.4 $\pm$ 0.23} & {\bf 24.5 $\pm$ 0.35}  & {\bf 46.3 $\pm$ 0.35} & {\bf 75.0 $\pm$ 1.70}\\
  SVGP (Time in s) & { 188 $\pm$ 1.18} & / & / & / & /\\
  \bottomrule
  \end{tabular}%
  }
\end{table*}

\section{EXPERIMENTS}

In the following, we evaluate AS-VGP on a number of regression tasks. We highlight the following properties of our method: 1) AS-VGP significantly reduces the memory requirements of sparse variational GPs without sacrificing on performance. 2) AS-VGP is extremely fast and scalable (training on 2 million 1D data points and 1000 inducing points in under 6 seconds). 3) AS-VGP is able to perform closed-form optimal variational inference when other methods have to use stochastic optimisation instead. 4) AS-VGP is not limited to low-dimensional problems and improves upon VFF when using an additive structure. 5) AS-VGP is particularly suited to modelling fast-varying spatial datasets. 

\subsection{One-Dimensional Regression} \label{sec:1D-regression}

\looseness=-1\textbf{Regression Benchmarks.}\quad The purpose of this experiment is to assess the empirical performance and computational benefits of AS-VGP in comparison with SVGP and VFF on medium-sized datasets. We use three UCI benchmarks and a synthetic dataset to compare the predictive performance of AS-VGP with SVGP and VFF. For the synthetic dataset, we generated a periodic function to compare our locally supported B-Spline basis with VFF, which uses a naturally periodic basis. 

For each dataset, we randomly sample 90$\%$ of the data for training and 10$\%$ for testing, repeating this five times to calculate the mean and standard deviation, of the predictive performance (MSE) and uncertainty quantification (NLPD). When using AS-VGP, we normalise the inputs to be between $[0,M]$, where $M$ is the number of inducing points to ensure the spacing between knots is equal to $1$, to avoid numerical issues caused by large gradients when computing the inner-product between basis functions. All models are trained using the L-BFGS optimiser; for VFF and AS-VGP, we precompute the matrix product $\Kuf\Kfu$. We use the Mat\'ern-3/2 kernel for each experiment.

The results in Table \ref{Tab:regression_benchmakrs} demonstrate that AS-VGP is comparative in performance to VFF on every dataset, whilst being less memory intensive. This highlights how our locally supported basis functions offer benefits in both complexity and memory over their globally supported counterparts, while retaining comparable performance. We note that SGPR performs slightly better than both VFF and AS-VGP, but at a higher computational complexity and without the ability for pre-computation.

\begin{figure}[t]
    \centering
    \includegraphics[width=\columnwidth]{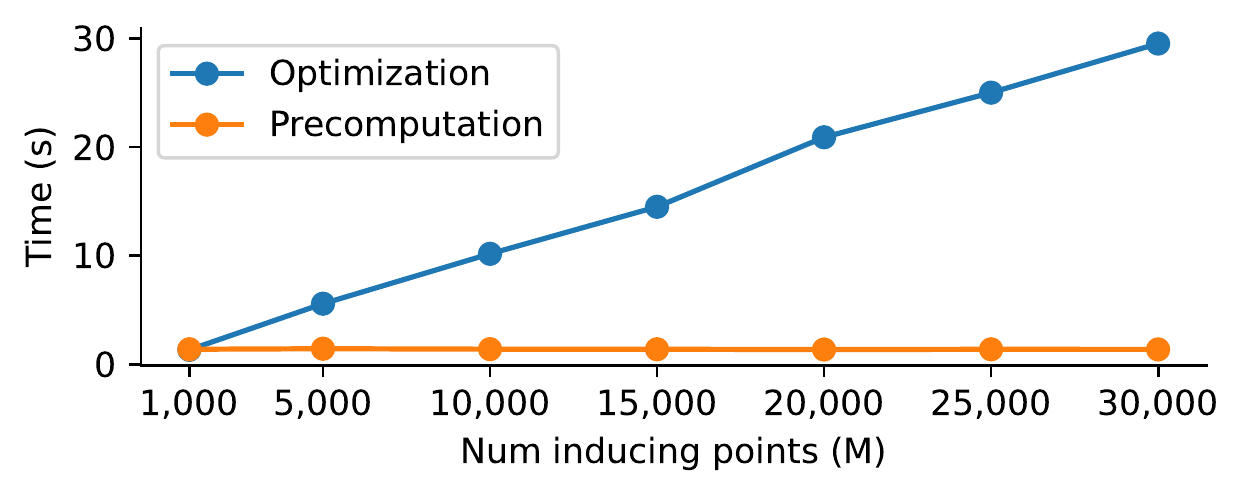}
    \label{fig:linear}
    \vspace*{-7mm}
    \caption{Illustration of the linear scaling in computation of the ELBO and independence on computing the product $\Kuf\Kfu$ w.r.t. the number of inducing points.}
\end{figure}

\textbf{Large-Scale Regression.} \quad In this example, we illustrate the scalability of our method both in the number of data points and in the number of inducing points, using the household electric power consumption dataset, where $N=2,049,279$. We opt to use the entire dataset, which uses a one-minute sampling rate over a period of four years, as an example of data with a very low lengthscale. This necessitates a large number of inducing variables and tests the model's ability to scale accordingly. We repeat each experiment five
times by randomly sampling 95\% of the data for training and
use the remaining 5\% for evaluation. For each experiment, we use the Matérn-3/2 kernel. 

Results are displayed in Table \ref{Tab:large_scale_regression}.%
We were unable to use either VFF or VISH in this experiment as we were unable to precompute $\Kuf\Kfu$ due to $\Kuf$ not fitting on GPU memory. For SVGP, we used minibatching to reduce the reliance on memory. However, we also ran into issues when using $M\ge1000$ requiring us to use a very ($N_b = 100$) small batchsize given the size of the dataset and the model became computationally unfeasible for $M\ge5,000$.
\looseness=-1In contrast, for AS-VGP, memory was not an issue, and we were able to efficiently scale the number of inducing points. As highlighted in Table \ref{Tab:large_scale_regression}, our method was more than two orders of magnitude faster than SVGP, fitting a GP with $1,000$ inducing points and over $2$ million datapoints in under $6$ seconds. We also observe that the time taken for each AS-VGP experiment follows a linear trend shown in Figure \ref{fig:linear} as predicted (see Table~\ref{Tab:SGPR_Complexity}).  
AS-VGP was also more accurate than SVGP both in predictive performance (MSE) and uncertainty quantification (NLPD), and showed an increase in performance as more inducing variables were added. Firstly, this is indicative of optimal closed-form variational inference being a better approximation to the true posterior than stochastic variational inference. Secondly, this emphasises how the B-spline basis is able to accurately represent local variance in the data  and motivates using a large number of inducing points when the lengthscale is very small. 

\subsection{Additive Regression}

\begin{figure*}[t]
\centering
\includegraphics[width=0.8\textwidth]{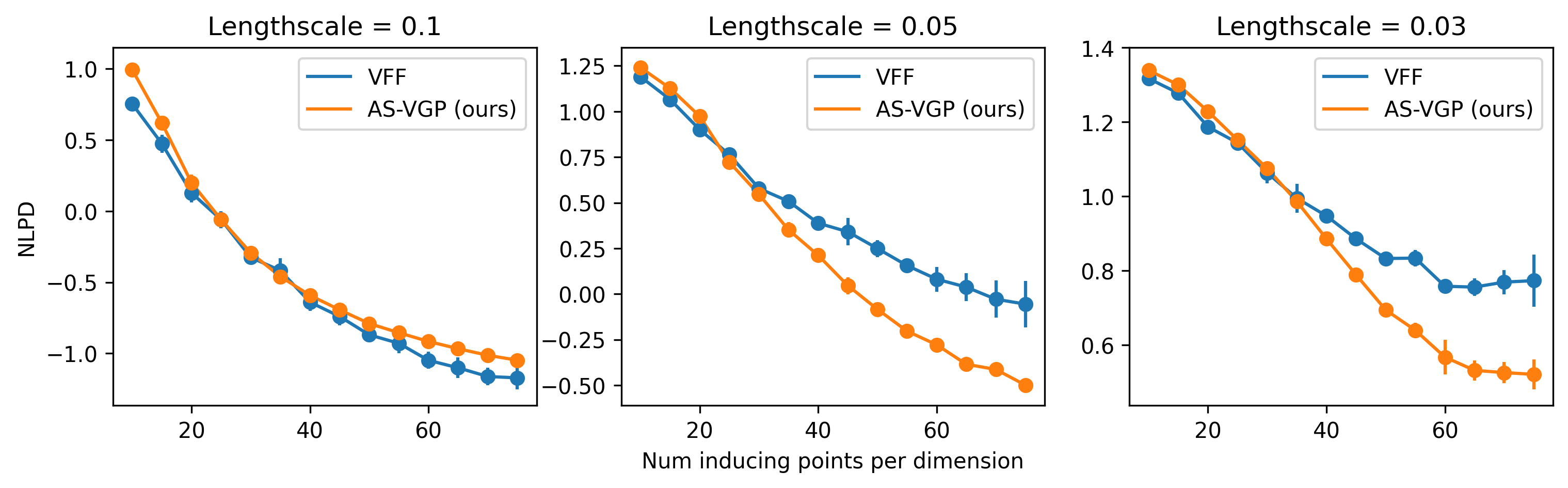}
\vspace*{-3mm}
\caption{Mean NLPD for AS-VGP and VFF for increasing numbers of inducing points. The data is obtained by sampling a GP with Mat\'ern-3/2 kernel with decreasing lengthscale. The mean NLPD for each model is computed with known parameters and by averaging over five separate samples. The error bars show one standard deviation.}
\label{fig:nlpd_spatial_data}
\end{figure*}

\begin{table*}
  \caption{Predictive mean squared errors (MSEs) and negative log predictive densities (NLPDs) with one standard deviation based on 5 random splits for a number of UCI regression datasets. All models use a Mat\'ern-3/2 kernel and the L-BFGS optimiser for training. All models show comparable performance.}
  \label{Tab:airline}
  \centering
  \resizebox{2\columnwidth}{!}{%
  \begin{tabular}{llllllllll}
  \toprule
  &&
  \multicolumn{2}{c}{$N=10,000$} &
  \multicolumn{2}{c}{$N=100,000$} &
  \multicolumn{2}{c}{$N=1,000,000$} &
  \multicolumn{2}{c}{$N=5,929,413$}\\ 
  \cmidrule(lr){3-4}
  \cmidrule(lr){5-6}
  \cmidrule(lr){7-8}
  \cmidrule(lr){9-10}
  {Model} & {M} &
  \hfil MSE & 
  \hfil NLPD & 
  \hfil MSE & 
  \hfil NLPD &
  \hfil MSE & 
  \hfil NLPD &
  \hfil MSE & 
  \hfil NLPD \\
  \midrule
  VISH &
  610 &
  0.90$\pm$0.16 &
  1.33$\pm$0.09 &
  0.81$\pm$0.05 &
  1.27$\pm$0.03 &
  0.83$\pm$0.03 &
  1.28$\pm$0.01 &
  0.83$\pm$0.06 &
  1.27$\pm$0.00 \\
  VFF &
  30/dim &
  0.89$\pm$0.15 &
  1.36$\pm$0.09 &
  0.82$\pm$0.05 &
  1.32$\pm$0.03 &
  0.83$\pm$0.01 &
  1.34$\pm$0.01 &
  0.83$\pm$0.00 &
  1.32$\pm$0.00 \\
  AS-VGP &
  30/dim &
  0.95$\pm$0.17 &
  1.39$\pm$0.09 &
  0.84$\pm$0.05 &
  1.33$\pm$0.03 &
  0.84$\pm$0.01 &
  1.33$\pm$0.01 &
  0.83$\pm$0.00 &
  1.33$\pm$0.00 \\
  AS-VGP &
  200/dim &
  0.91$\pm$0.16 &
  1.37$\pm$0.09 &
  0.82$\pm$0.05 &
  1.32$\pm$0.03 &
  0.83$\pm$0.01 &
  1.32$\pm$0.01 &
  0.82$\pm$0.00 &
  1.32$\pm$0.00
  \\
  \bottomrule
  \end{tabular}%
  }
\end{table*}

In this experiment we show that AS-VGP is not limited to low-dimensional problems, but can scale to high dimensions using an additive structure.

\looseness=-1The airline dataset is a common GP benchmark, consisting of flight details for every commercial flight in the USA from 2008. The task is to predict the amount of delay $y$ given eight different covariates (route distance, airtime, aircraft, age, etc.). We follow the exact same setup as from \cite{hensman2013gaussian} using an additive Mat\'ern-3/2 GP and evaluate the performance on four datasets of size $10K$, $100K$, $1,000K$ and $5,929,413$ (complete dataset) by subsampling the original data. For each dataset, we perform 10 splits, using two thirds of the data for training and a third for testing. We report the mean and standard deviation of the MSE and NLPD in Table \ref{Tab:airline}. For AS-VGP, we normalise the inputs to be between $[0,M]$, where $M$ is the number of inducing points to ensure the spacing between knots is equal to $1$. 

We compare our method using both $M=30$ ($240$ in total) basis functions and $M=200$ ($1600$ in total) basis functions per dimension. Table \ref{Tab:airline} shows that by adding more basis functions, we can improve upon VFF and VISH in terms of MSE on the larger datasets. We can also scale to larger numbers of inducing points. While VFF uses 240 inducing points in total, we use 1600 in our largest experiment, while remaining computationally efficient since pre-computation is independent of the number of inducing points (and linear in the number of training datapoints; see Table~\ref{Tab:SGPR_Complexity}).

\subsection{Synthetic Spatial Data}
\looseness=-1In the following, we demonstrate the effectiveness of AS-VGP on synthetic spatial data with an inherently low lengthscale. To simulate high fidelity spatial data with fast variations, we sample from a 2D GP with a kernel constructed as the product of two 1D Mat\'ern-3/2 kernels. We generate data by sampling the GP five times each for three different lengthscales: $0.1$, $0.05$ and $0.03$. Fixing the lengthscales of AS-VGP and VFF to match the generated data, we then compute the NLPD, for different numbers of inducing points.

Figure \ref{fig:nlpd_spatial_data} shows that AS-VGP captures the variance in the data better than VFF as the lengthscale is reduced. This is in part the fault of the product basis in VFF, which produces features with very small variance, becoming more pronounced with features of higher-frequency (\cite{dutordoir2020sparse}).
However, it also promotes the use of compactly supported basis functions which, unlike the Fourier basis that describe the process across the entire domain, act locally and therefore are more effective at modelling local variations in the data. 

\subsection{Real-World Spatial Data}
\begin{figure*}[h]
\centering
\subfigure[Ground truth.]{
\includegraphics[width=0.3\hsize]{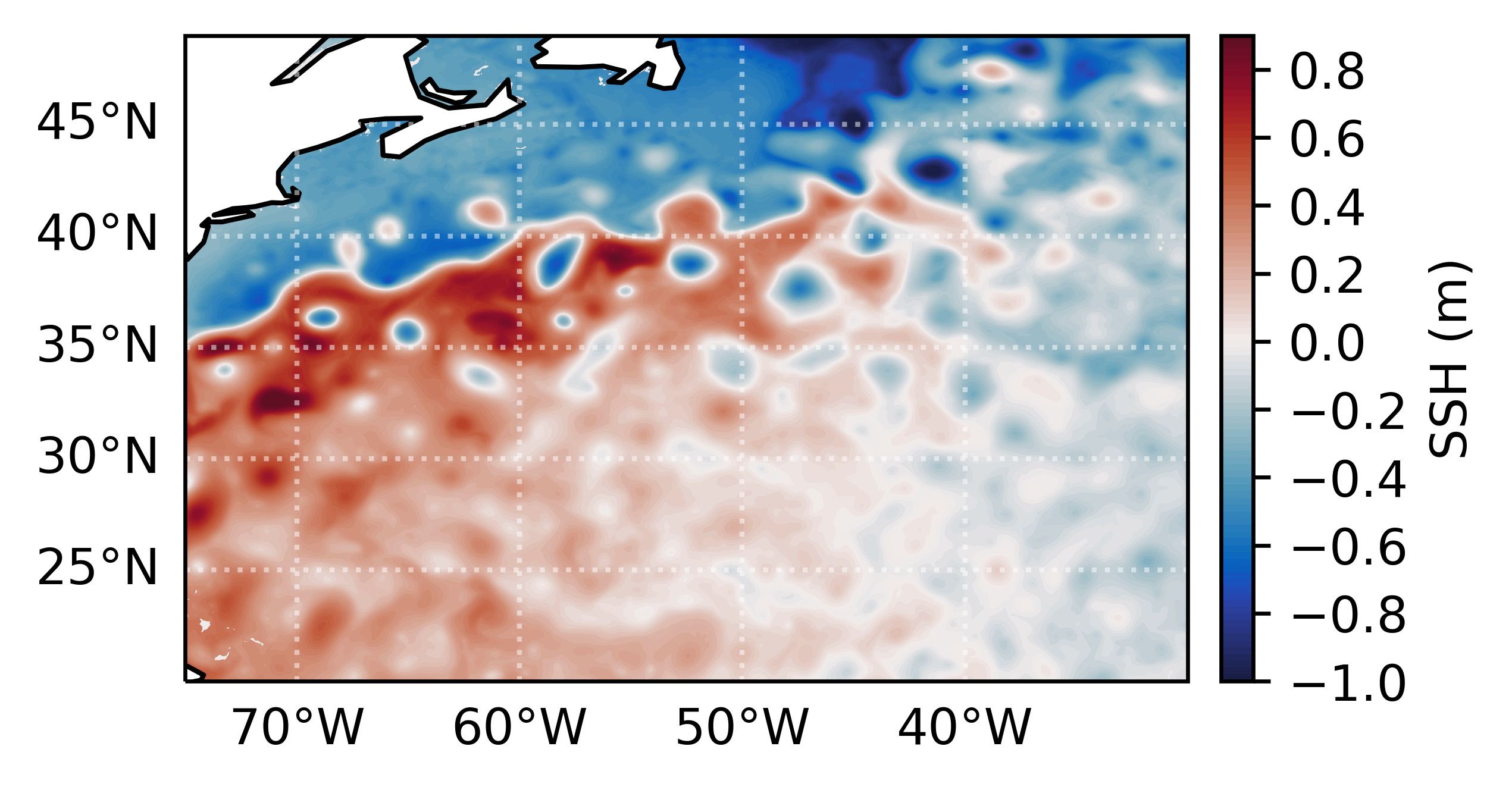}
\label{fig:ground truth}
}
\subfigure[Predictive mean.]{
\includegraphics[width=0.3\hsize]{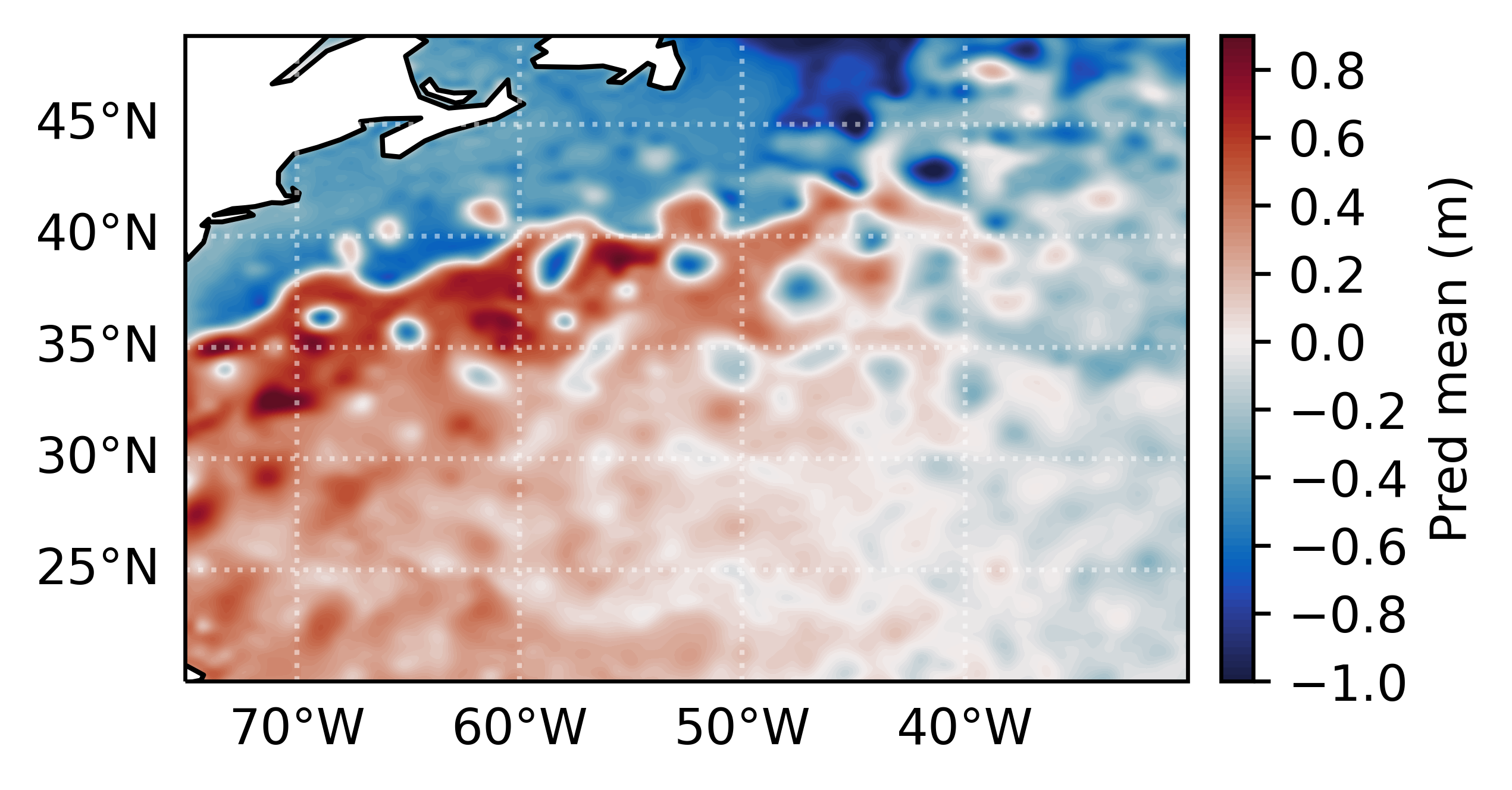}
\label{fig:pred mean}
}
\subfigure[Predictive standard deviation.]{
\includegraphics[width=0.3\hsize]{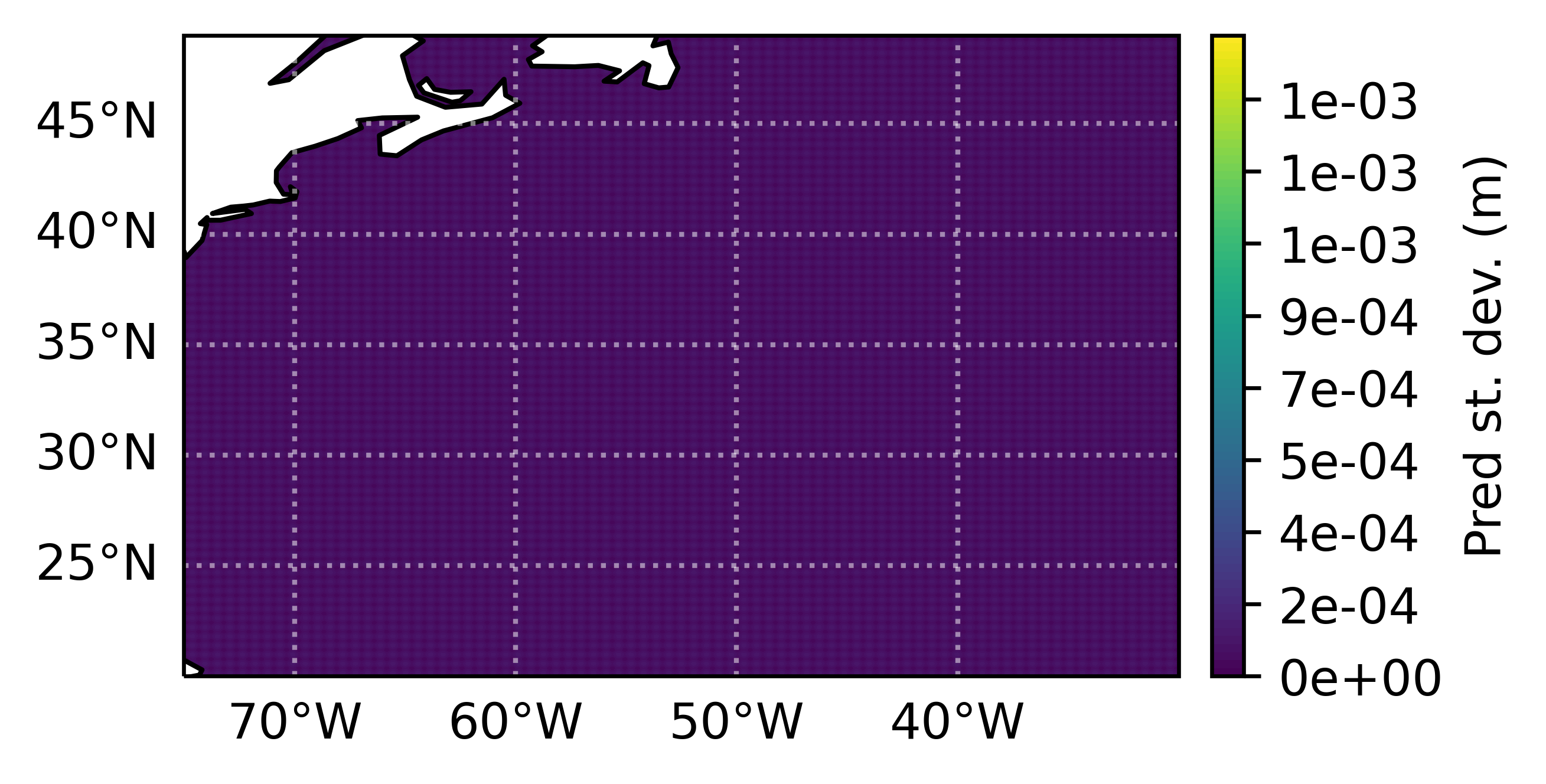}
\label{fig:pred std}
}
\caption{Real-world data from the eNATL60 ocean model over the Gulfstream at $1/60^\circ$ grid resolution. \subref{fig:ground truth} Ground truth; \subref{fig:pred mean} Predictive mean and \subref{fig:pred std} predictive standard deviation for AS-VGP at a regular grid with $1/12^\circ$ resolution. The predictive mean of the AS-VGP and the ground truth are nearly identical while the predictive uncertainty essentially vanishes.}
\label{fig:eNATL60}
\end{figure*}

In this experiment, we test AS-VGP on a very large spatial regression problem. For this we use the eNATL60 ocean model of sea surface height (SSH) over the North Atlantic at $1 / 60^{\circ}$ grid resolution as a real-world example of an extremely large low-lengthscale spatial regression. We perform a typical regridding problem by interpolating the model data defined on a curvilinear grid onto a regular latitude-longitude grid. We restrict the domain to a $45^\circ \times 30^\circ$ region and randomly select 2 million data points from the model as training observations and $100,000$ points for testing. We then evaluate the trained AS-VGP model on a regular grid at $1/12^\circ$ resolution, equivalent to a $540\times 360$ grid.

\looseness=-1 We fit AS-VGP on this data using 100 basis function per dimension (10,000 in total) in 109 seconds, 41 seconds for pre-computation and 68 seconds for optimisation, achieving an MSE of $9.3 \times 10^{-4}$ and NLPD of $2.1$ on the test set. Similar to the 1D large-scale regression experiment, we could not use the equivalent VFF or SGPR model as storing $\Kuf$ (a $2,000,000 \times 10,000$ matrix), requires 149 GB of memory, which cannot be stored on GPU or CPU memory. In contrast, for AS-VGP, storing $\Kuf$ only requires $216$ MB of memory. Consequently, AS-VGP can handle both large numbers of datapoints and inducing points without requiring stochastic optimsation, which is unachievable using both SGPR and VFF. Using the trained model, we then predict onto a regular latitude longitude grid at $1/ 12^\circ$ resolution, taking 36 seconds. Figure \ref{fig:eNATL60} shows that AS-VGP is able to model the small structures present in the SSH, quickly and efficiently, whilst still being able to perform closed-form optimal variational inference where other methods can't.

\section{DISCUSSION}
For one-dimensional inputs with a Gaussian likelihood, AS-VGP is extremely fast, scalable and lightweight, exploiting the band-diagonal structure of both $\Kuu$ and $\Kuf\Kfu$ to perform pre-computation that scales linearly in the number of datapoints and evaluations of the ELBO that scale linearly in the number of inducing variables. We also show that our method is not limited to one-dimensional inputs, but can scale to higher dimensions using an additive or Kronecker structure. In particular, we show that our method is particularly strong at representing processes with small lengthscales, making it amenable to modelling spatio-temporal data or long time series.

Stochastic variational inference also enables the scaling of GPs to large datasets via mini-batching. In practice, however, when using a Gaussian likelihood and if compute permits, SGPR produces a better approximation to the true posterior than SVGP. Whilst in regular SGPR the cost to compute the ELBO is dependent on $N$, VFF made it possible to remove this by performing a one-off pre-computation, effectively scaling SGPR to millions of data points. Our work extends VFF even further by reducing the complexity of the pre-computation with respect to the number of datapoints and decoupling it from the number of inducing variables, enabling us to scale to larger $N$ and larger $M$. 

Comparisons to our method can also be made to Structured Kernel Interpolation (SKI) by \cite{wilson2015kernel}, but from a variational perspective. Both SKI and AS-VGP construct inducing variables on dense grids. However, whereas SKI performs explicit interpolation between inducing points, AS-VGP implicitly performs interpolation by instead constructing inducing variables equivalent to evaluating a set of B-spline basis functions. 

Whilst, we show good performance in low-dimensional problems, ideally, we would not have to impose a Kronecker structure that scales so badly in dimensionality, but instead project directly onto a set of 2D basis functions. Taking inspiration from the connections with SKI, a better choice of basis might be one defined on the simplex, which offers linear scaling in $D$ when generalising to higher dimensions \citep{kapoor2021skiing}.

\textbf{Limitations} \quad The main limitation of our approach is the scaling to high dimensions. Unlike VISH, we inherit many of the shortcoming of VFF, including a reliance on tensor products which requires an exponential increase in the number of basis functions with increasing dimensions. However, by decoupling the pre-computation from the number of inducing variables, our method is less affected by exponential scaling than VFF. For low numbers of inducing features $M$, our method performs worse than VFF. However, we can mitigate this shortcoming by using more inducing features due to the linear scaling in $M$ versus cubic for VFF (see Table \ref{Tab:SGPR_Complexity}). Finally, like VFF our method currently only supports the Mat\'ern class of kernels. A future research direction would be to expand the class of kernels that can be decomposed using B-splines, e.g., non-stationary kernels, which could help improve spatial modelling.   

In practise we propose to use our method on low-dimensional problems ($D\le 4$), such as spatial or spatio-temporal data, where our method has shown to be computationally and memory efficient while being able to  capture high-frequency variations. 

\section{CONCLUSION}
We introduced a novel inter-domain GP model wherein the inducing features are defined as RKHS projections of the GP onto compactly-supported B-spline basis functions. This results in covariance matrices that are sparse, allowing us to draw entirely on techniques from sparse linear algebra to do GP training and inference and thereby opening the door to GPs with tens of thousand inducing variables. Our experiments demonstrate that we get significant computational speed up and memory savings without sacrificing accuracy. 

\section*{Acknowledgements}
HJC is supported by the UCL Department of Computer Science DTP scholarship.

\bibliography{main.bib}

\appendix
\onecolumn

\section{CODE}

Code is available at \url{https://github.com/HJakeCunningham/ASVGP}

\section{EXPERIMENT DETAILS} \label{app:experiments}

The timed experiments (Table \ref{Tab:large_scale_regression}) were performed using a AMD Ryzen 2920X 12-Core CPU and an NVIDIA GeForce RTX 2080 GPU. Below, we include specific details on the two experiments conducted.

\paragraph{Regression Benchmarks}
For the synthetic dataset, we generate 10,000 random noisy observations from the test function
\begin{equation*}
    f(x) = \mathrm{sin}(3\pi x) + 0.3\cos(9\pi x) + \frac{\sin(7\pi x)}{2}.
\end{equation*}

\paragraph{Metrics} 
We use the mean-squared error (MSE) and the negative log-predictive density (NLPD) to evaluate the performance of our model. These are defined as
\begin{align}
    \text{MSE}(\{X_n, y_n\}_{n=1}^N) &= \frac{1}{N} \sum_{n=1}^N \|y_n - \mu(X_n)\|^2, \\
    \begin{split}
    \text{NLPD}(\{X_n, y_n\}_{n=1}^N) &= -\frac{1}{N} \sum_{n=1}^N\mathrm{log} \int p(y_n | f_n) \,\mathcal{N}(f_n | \mu(X_n), \xi(X_n)) \,\mathrm{d} f_n,
    \end{split}
\end{align}
where $\mu, \xi$ are the posterior mean and variance, respectively.

\section{RKHS INNER PRODUCTS}

The inner products corresponding to the Mat\'ern-1/2 and Mat\'ern-3/2 RKHS defined over the domain $\mathcal{D}=[a,b]$, as given by \cite{durrande2016detecting, hensman2017variational},  are
\begin{align}
    \langle f, g \rangle_{\mathcal{H}_{k_{1/2}}} &= \frac{l}{2\sigma^2}\int_a^b f' g' \diff x + \frac{1}{2l\sigma^2} \int_a^b fg \,\diff x + \frac{1}{2\sigma^2}[f(a)g(a)+f(b)g(b)], \\
    \addlinespace
    \begin{split}
    \langle f, g \rangle_{\mathcal{H}_{k_{3/2}}} &= \frac{l^3}{12\sqrt{3}\sigma^2}\int_a^b f'' g'' \diff x + \frac{l}{2\sqrt{3}\sigma^2} \int_a^b f'g' \diff x + \frac{\sqrt{3}}{4l\sigma^2}\int_a^b fg \,\diff x \\
    &\quad + \frac{1}{2\sigma^2}[f(a)g(a)+f(b)g(b)] + \frac{l^2}{2\sigma^2}[f'(a)g'(a)+f'(b)g'(b)],
    \end{split}
\end{align}
respectively, where $l, \sigma$ are the lengthscale and amplitude hyperparameters. 

When performing RKHS projections, it is important to note that the B-splines must belong to the RKHS defined by our choice of kernel. As stated by \cite{kanagawa2018gaussian}, the RKHS generated by the Mat\'ern-$\nu/2$ kernel is norm-equivalent to the Sobolev space $\mathcal{H}^{\nu/2+1/2}$. Due to their piecewise polynomial form, B-splines of order $k$ can be shown to belong to the Sobolev space $\mathcal{H}^{k}$ (see Section \ref{app:splines}). 

From Section \ref{Actually Sparse Variational GP Inference}, to minimise computational complexity, we wish to use B-spline basis functions with minimal bandwidth of the $\Kuu$ matrix. As a result, for the Mat\'ern-$\nu/2$ kernel we project onto B-splines of order $\nu/2+1/2$.

\section{GENERALISED GAUSSIAN FIELDS OVER RKHS}
\label{app:generalised-gaussian-fields}
In this appendix, we justify our abuse of notation $\left<f, \phi\right>_{\mathcal{H}}$ when $f$ is a GP whose RKHS is $\mathcal{H}$. To this end, we first introduce the notion of a generalised Gaussian field as follows.

\begin{defn}[\cite{lototsky2017stochastic}]
Let $\mathcal{H}$ be a Hilbert space and $(\Omega, \mathcal{F}, \mathbb{P})$ a probability space. We say that a function $F : \Omega \times \mathcal{H} \rightarrow \mathbb{R}$ is a zero-mean generalised Gaussian field over $\mathcal{H}$ if the random variable $Fh := F(\cdot, h)$ is Gaussian for all $h \in \mathcal{H}$ and the following property holds:
\begin{enumerate}
    \item $\mathbb{E}\left[Fh\right] = 0$ for all $h \in \mathcal{H}$, and
    \item $\mathrm{Cov}\left[Fg, Fh\right] = \left<g, h\right>_\mathcal{H}$ for all $g, h \in \mathcal{H}$, where $\left<\cdot, \cdot\right>_\mathcal{H} : \mathcal{H} \times \mathcal{H} \rightarrow \mathbb{R}$ denotes the inner product on $\mathcal{H}$.
\end{enumerate}
\end{defn}

In the special case when $\mathcal{H}$ is an RKHS over a base space $X$ with kernel $k : X \times X \rightarrow \mathbb{R}$, we can further identify generalised Gaussian fields $F : \Omega \times \mathcal{H} \rightarrow \mathbb{R}$ with a stochastic process $f : \Omega \times X \rightarrow \mathbb{R}$ over $X$ by the following relation:
\begin{align}
    f(x) = Fk(\cdot, x), \quad \text{for all} \quad x \in X. \label{eq:identify-gp-samples-with-generalised-field}
\end{align}
Notice that $k(\cdot, x) \in \mathcal{H}$ so the expression on the RHS of \eqref{eq:identify-gp-samples-with-generalised-field} makes sense.
Indeed, GPs of the Matérn class can be identified with generalised Gaussian field over the Sobolev space \cite{borovitskiy2020matern}. Hence, for any Matérn-$\nu/2$ GP $f$ and any $\phi \in \mathcal{H} := \mathcal{H}^{\nu/2+1/2}$, we can define the RKHS projection of $f$ onto $\phi$ by
\begin{align}
    \left<f, \phi\right>_\mathcal{H} := F\phi,
\end{align}
where $F$ is the generalised Gaussian field corresponding to the process $f$.

\section{IMPLEMENTATION OF B-SPLINES}
\label{app:splines}
The $m$-th B-spline basis function of order $k$, which we denote by $B_{m,k}(x)$ can be computed according to the Cox-de-Boor recursion formula \citep{prautzsch2002bezier}
\begin{align}
    B_{m, 0}(x) &= \begin{cases}
    1, & \text{if $v_m\le x \le v_{m+1}$},\\
    0, & \text{otherwise}, 
  \end{cases} \\
  \addlinespace
  \begin{split}
  B_{m, k}(x) &= \frac{x-v_m}{v_{m+k} - v_m} B_{m, k-1}(x) + \frac{v_{m+k+1} - x}{v_{m+k+1} - v_{m+1}} B_{m+1, k-1}(x).
  \end{split}
\end{align}
The case $k=0$ corresponds to a top-hat function $\mathbf{1}_{[v_m, v_{m+1}]}(x)$, with support spanning a single sub-interval. In the case $k=1$, we have the piecewise linear function
\begin{align}
    B_{m, 1}(x) &= 
    \begin{dcases}
    \frac{x-v_m}{v_{m+1}-v_m}, \quad & \text{for } x \in [v_m, v_{m+1}], \\
    \frac{v_{m+2}-x}{v_{m+2}-v_{m+1}}, \quad & \text{for } x \in [v_m, v_{m+1}], \\
    0, \quad &\text{otherwise},
    \end{dcases}
\end{align}
which corresponds to the tent map, spanning two sub-intervals (see Figure \ref{fig:splines} (a)). In the case $k=2$, we have the piecewise quadradic function
\begin{align}
B_{m,2}(x) = 
\begin{dcases}
    \frac{(x-v_m)^2}{(v_{m+2}-v_m)(v_{m+1}-v_m)}, \quad & \text{for } x \in [v_m, v_{m+1}], \\
    \frac{(x-v_m)(v_{m+2}-x)}{(v_{m+2}-v_m)(v_{m+2}-v_{m+1})} + \frac{(v_{m+3}-x)(x-v_{m+1})}{(v_{m+3}-v_{m+1})(v_{m+2}-v_{m+1})}, \quad & \text{for } x \in [v_{m+1}, v_{m+2}], \\
    \frac{(v_{m+3}-x)^2}{(v_{m+3}-v_{m+1})(v_{m+3}-v_{m+2})}, \quad & \text{for } x \in [v_{m+2}, v_{m+3}], \\
    0, \quad & \text{otherwise},
\end{dcases}
\end{align}
spanning three sub-intervals (see Figure \ref{fig:splines} (b)). Likewise, we can construct a piecewise cubic polynomial corresponding to the case $k=3$ (Figure \ref{fig:splines} (c)).

We claim that the B-spline $B_{m,k}(x)$ is of class $C^{k-1}$ for $k \geq 2$ and is moreover $k$-times weakly differentiable for $k \geq 1$. To prove the first claim, when $k \geq 2$, one can show that \citep{prautzsch2002bezier}
\begin{align}
\begin{split}
    \frac{\diff B_{m,k}(x)}{\diff x} &= \frac{k}{v_{m+k} - v_m} B_{m, k-1}(x) - \frac{k}{v_{m+k+1} - v_{m+1}} B_{m+1, k-1}(x).
\end{split}
\end{align}
Thus, the first derivative of $B_{m, k}(x)$ is a linear combination of B-splines of order $k-1$, its second derivative is a linear combination of B-splines of order $k-2$, and so on. Now, since B-splines of order $k-i$ are continuous in $x$ if and only if $k-i \geq 1$, we see that the $i$-th derivative of $B_{m, k}(x)$ is continuous in $x$ provided $i \in \{1, \ldots, k-1\}$. This implies that $B_{m, k}(x)$ is of class $C^{k-1}$.

Next, we show that $B_{m,k}(x)$ is $k$-times weakly differentiable. In the case $k \geq 2$, by the previous arguments, we know that its $k-1$-th derivative exists and moreover can be expressed as a linear combination of first order B-splines. Hence, all we need to demonstrate is that the weak derivative of first order B-splines exists. This can be shown easily, with
\begin{align}
    D_x^w B_{m, 1}(x) =
    \begin{dcases}
    \frac{1}{v_{m+1}-v_m}, \quad & \text{for } x \in [v_m, v_{m+1}], \\
    \frac{1}{v_{m+1}-v_{m+2}}, \quad & \text{for } x \in [v_m, v_{m+1}], \\
    0, \quad &\text{otherwise},
    \end{dcases}
\end{align}
where $D^w_x$ denotes the weak derivative with respect to $x$. This also trivially implies the case $k=1$.
Hence, $B_{m,k}(x)$ is $k$-times weakly differentiable for all $k \geq 1$.

\end{document}